\pdfoutput=1
\documentclass[11pt]{article}

\usepackage[]{naacl2021}
\usepackage{times}
\usepackage{latexsym}

\usepackage[T1]{fontenc}
\usepackage[utf8]{inputenc}

\usepackage{microtype}

    \usepackage{xcolor}
    \usepackage{lipsum}
    \usepackage{graphicx}
    \usepackage{amsmath}
    \usepackage{amssymb}
    \usepackage[capitalise]{cleveref}
    
    \usepackage[acronym]{glossaries}
    \newacronym{sst}{SST}{Stanford Sentiment Treebank}
    \newacronym{sp}{SP}{Submodular Pick}
    \newacronym{arct}{ARCT}{Argument and Reasoning Comprehension Task}

    \newcommand{\mpara}[1]{\medskip\noindent{\bf #1}}

\title{Towards Benchmarking the Utility of Explanations for Model Debugging}
\author{Maximilian Idahl\textsuperscript{1} \quad Lijun Lyu\textsuperscript{1}  \quad Ujwal Gadiraju\textsuperscript{2} \quad Avishek Anand\textsuperscript{1}\\
  \textsuperscript{1}L3S Research Center, Leibniz University of Hannover / Hannover, Germany\\
  \textsuperscript{2}Delft University of Technology / Delft, Netherlands\\
  {\texttt{\{idahl, lyu, anand\}@l3s.de} \quad \texttt{u.k.gadiraju@tudelft.nl}}
}

\begin{document}
\maketitle
\begin{abstract}
    Post-hoc explanation methods are an important class of approaches that help understand the rationale underlying a trained model's decision.
    But how useful are they for an end-user towards accomplishing a given task?
    In this vision paper, we argue the need for a benchmark to facilitate evaluations of the utility of post-hoc explanation methods.
    As a first step to this end, we enumerate desirable properties that such a benchmark should possess for the task of debugging text classifiers.
    Additionally, we highlight that such a benchmark facilitates not only assessing the effectiveness of explanations but also their efficiency.
\end{abstract}

\section{Introduction}
A large variety of post-hoc explanation methods have been proposed to provide insights into the reasons behind predictions of complex machine learning models~\cite{ribeiro2016LIME,sundararajan2017IntegratedGrads}.
Recent work on explainable machine learning in deployment~\cite{bhatt2020ExplainableMachineLearninginDeployment} highlights that explanations are mostly utilized by engineers and scientists to debug models.

The use of explanations for model debugging is motivated by their ability to help detect \emph{right for the wrong reasons} bugs in models.
These bugs are difficult to identify from observing predictions and raw data alone and are also not captured by common performance metrics computed on i.i.d. datasets.
Deep neural networks are particularly vulnerable to learning decision rules that are right for the wrong reasons.
They tend to solve datasets in unintended ways by performing shortcut learning~\cite{geirhos2020shortcut}, picking up spurious correlations, which can result in ``Clever Hans behavior''~\cite{lapuschkin2019unmasking}.
Considering this important role of explanations during the model validation or selection phase, we call for more utility-focused evaluations of explanation methods for model debugging.

We identify two key limitations in current approaches for measuring the utility of explanations for debugging:
1) A ground-truth problem, and 2) an efficiency problem.

% Explanations are used to perform a sanity check after model training, aiming to verify whether the model's behavior aligns with human (expert) intuition on various data points.
% The goal is to avoid deploying models that are \textit{right for the wrong reasons}, which are difficult to identify from observing predictions and raw data only.

% While the evaluation of post-hoc explanations has mostly focused on measuring how well they help humans understand the behavior of complex models ("simulation") and how much they improve human decision making in human-machine collaboration setups, assessing the utility of explanations for debugging purposes has fallen short, especially in the language domain.

First, in all common evaluation setups, the presence of bugs serves as a \emph{ground truth} and although crucial to the evaluation's outcome, intentionally adding bugs to create models that exhibit \emph{right for the wrong reasons} behavior has not been thoroughly studied.
We envision a benchmark collection of verified buggy models to encourage comparable utility-centric evaluations of different explanation methods.
Bugs can be injected into models by introducing artificial decision rules, so-called \emph{decoys}, into existing datasets.
To establish a rigorous design of decoy datasets, we enumerate desirable properties of decoys for text classification tasks.
While a decoy has to be \emph{adoptable} enough to be verifiably picked up during model training, the resulting decoy dataset should also be \emph{natural}.

Second, the utility of explanations is not only determined by their \emph{effectiveness}.
For local explanation methods, i.e.,~methods that generate explanations for individual instances, the selection of instances examined by humans is crucial to the utility of explanation methods, and thus successful debugging.
This \emph{efficiency} problem of \emph{how fast users can detect a bug} has been mostly ignored in previous evaluations. By presenting only instances containing a bug they implicitly assume the selection process to be optimal; an assumption that does not transfer to real-world scenarios and potentially leads to unrealistic expectations regarding the utility of explanations.

% We begin by outlining the evaluation approaches taken by previous works to measure how useful explanations are for model debugging in Section~\ref{sec:eval_setups}.
% Next, we discuss the intentional creation of buggy models via decoy datasets to alleviate the ground truth problem (Section \ref{sec:groundtruth_problem}), and establish properties and desiderata of decoy datasets for language tasks (Section \ref{sec:decoys_for_language}).
% Then, we highlight the efficiency problem posed by selecting instances for inspection in Section \ref{sec:efficiency_problem}.
% We conclude by providing an outlook how a benchmark collection of decoy datasets and corresponding buggy models could greatly improve the utility-centric evaluation of explanations, both in terms of effectiveness and efficiency.

\section{Evaluating the Utility of Explanations for Debugging}
\label{sec:eval_setups}

The utility of explanations is measured by \emph{how useful the explanation is to an end-user towards accomplishing a given task}.
In this work, we focus on the model developer (as the stakeholder). We outline four different task setups used in previous work.

\subsection{Setup I: Identify and Trust}
\label{subsec:distrust-identify}

In a first setting employed by \citet{ribeiro2016LIME} to evaluate whether explanations lead to insights, users are presented with the predictions as well as the explanations generated for a model containing a (known) bug.
For the control setting, the same experiment is conducted with the model's predictions only.
The utility of an explanation is measured by how well the explanation can help users to accurately \emph{identify} the \emph{wrong reasons} behind the model's decision making and whether they would \emph{trust} the model to make good predictions in the real world or not.

\subsection{Setup II: Model Comparison}
\label{subsec:selection-explanation}

In another setup used by \citet{ribeiro2016LIME} the explanations for two models with similar validation performance are presented to human subjects, but with a bug contained in only one of the models.
Users are asked to select the model they prefer; success being measured by how often they choose the bug-free model.

\subsection{Setup III: Identify and Improve}
\label{subsec:improve-w-explanation}

Similar to Setup I, users are shown predictions and explanations for a model that contains at least one bug.
Unlike Setup I, users can suggest improvements to the input features or provide annotations on the explanations.
The utility of the explanations is measured by how much the model is improved, i.e. the difference in test performance before and after debugging.
Improvements can be applied by retraining and either removing input features~\cite{ribeiro2016LIME} or integrating explanation annotations into the objective function via explanation regularization~\cite{ross2017RRR, liu2019PriorsOnFeatAttr, Rieger2020InterpretationsAreUseful}. 
Alternatively, features can also be disabled on the representation level~\cite{lertvittayakumjorn2020find}.
        
\subsection{Setup IV: Data Contamination}
\label{subsec: data-contamination}    
    
In a setup aimed at evaluating explanation-by-example methods, the training data itself is modified, such that a selected fraction of instances contains a bug that is then inherited by a model. 
For example, \cite{koh2017InfluenceFunctions} flip the labels of 10\% of the training instances to show that influence functions can help uncover these instances.
Here, the utility of the explanations is measured by how many of these instances were uncovered, and by the performance gain obtained by re-labeling the uncovered instances.

% \section{Purposely creating models that are right for the wrong reasons}

\section{Ground Truth for Debugging with Explanations}
\label{sec:groundtruth_problem}

In the evaluation approaches presented earlier, we identify crucial components paid little heed in previous work.
All the evaluation setups require a model containing one or multiple bugs.
The presence of these bugs serves as a \emph{ground truth} and thus they are crucial to the evaluation's outcome.

The bugs introduced into models in the evaluation regimes are ``well understood'' and added purposely. 
From the literature, these purposefully introduced artifacts are also known as \emph{decoys}.
Although crucial to the evaluation's outcome, these decoys have not been thoroughly studied.
% Decoy datasets are a popular tool to inject bugs into models.
As a first step towards a more rigorous design of decoy datasets, we define properties and desiderata for text classification tasks.
The use of explanations for the model debugging task is motivated by their ability to help detect \emph{right for the wrong reasons} bugs in models, and thus decoys should be designed accordingly.

\subsection{Decoy Datasets}% for Text Classification}
% \label{sec:decoy-text-classfication}
\label{sec:decoy-dataset}

% Decoy datasets serve the purpose of injecting bugs into models.
% While bugs can be introduced into models through other means, for example by directly contaminating the model's weights \cite{adebayo2020debugging_tests}, decoy datasets are particularly suited for injecting bugs that make the resulting model's predictions \textit{right for the wrong reasons}.
% In contrast, the model contamination bugs introduced by \citet{adebayo2020debugging_tests} result in the predictions of a model being wrong, and for detecting such bugs the use of explanations is most likely overkill.
Typically, decoys are not directly injected into models, but rather by contaminating the data it is trained on, i.e.,~by creating a \emph{decoy dataset}.
While bugs can be introduced into models through other means, for example by directly contaminating the model's weights \cite{adebayo2020debugging_tests}, decoy datasets are particularly suited for injecting bugs that make the resulting model's predictions \emph{right for the wrong reasons}.
In contrast, the model contamination bugs introduced by \citet{adebayo2020debugging_tests} result in the predictions of a model being wrong, and for detecting such bugs monitoring loss and standard performance metrics is sufficient. %the use of explanations is most likely overkill.

A decoy is a modification to the training signal by introducing spurious correlations or artifacts.
For example, \citet{ross2017RRR} used Decoy-MNIST, a modified version of MNIST \cite{lecun2010mnist} where images contain gray-scale squares whose shades are a function of the target label.
Similarly, \citet{Rieger2020InterpretationsAreUseful} create decoy variants of the \gls{sst} dataset \cite{socher2013sst} by injecting confounder words.
Both works use the decoy datasets to evaluate whether their proposed explanation regularizers can correct a model's wrong reasons towards the indented decision-making behavior.
To assess the utility of explanations for debugging, \cite{adebayo2020debugging_tests} use a decoy birds-vs-dogs image classification dataset by placing all birds onto a sky background and all dogs onto a bamboo forest background.

\subsection{Verifying Decoy Adoption}
\label{sec:verifying}

When using decoys, an important step is to verify if a model trained on a decoy dataset indeed ``adopts'' or learns a decoy.
Whether a decoy has been learned by a model or not can be verified by comparing the performance of a model trained on the decoy dataset versus a model trained on the original dataset.
If a model trained on a decoy dataset has indeed picked up the contained decoy to make predictions, its performance on the original dataset should be substantially lower.
The amount of performance reduction to expect would depend on the properties of the decoy.

\section{Properties of Decoys for Text Classification}
\label{sec:decoys_for_language}

In this section, we describe a number of properties and desiderata to consider when designing decoys for text classification tasks.

\citet{niven2019probing} analyze the nature of spurious statistical unigram and bigram cues contained in the warrants of the \gls{arct}~\cite{habernal2018ARCT} using three key properties, which we modify for describing token-based decoys in text classification datasets:

Let \(X\) be a dataset of labeled instances \((x_i, y_i)\) and \(X^d \subseteq X\) be the subset of instances containing a decoy~\(d\).
The \emph{applicability} \(a\) of a decoy \(d\) describes the number of instances affected by the decoy, that is,~\(a_d = |X^d|\). 
A decoy's \emph{productivity} \(p_d\) measures the potential benefit to solving the task by exploiting it.
We define it as the largest proportion of the decoy co-occurring with a certain class label for instances in \(X^d\):
\begin{equation}
    p_d = 
    \frac{
        \max\limits_{c \in C}
            % \Bigg(\sum_{i=1}^{a_d}
            \Bigg(\sum\limits_{y_j \in Y^d}
                \begin{cases}
                    1,& \text{if }y_j = c \\
                    0,& \text{otherwise}
                \end{cases}
            \Bigg)
    }
    {a_d}
\end{equation}
where \(C\) is the set of classes and \(Y^d\) the labels corresponding to instances in \(X^d\).

Finally, the signal strength provided by a decoy is measured by its \emph{coverage} \(c_d\). It is defined as the fraction of instances containing the decoy over the total number of instances:~\(c_d = a_d/|X|\).

\begin{figure*}[t]
    \includegraphics[width=\textwidth]{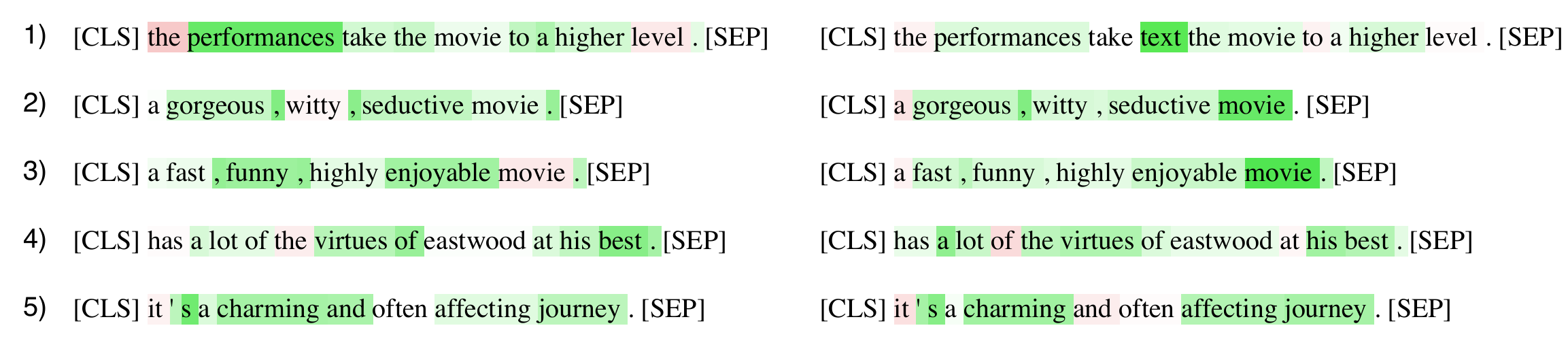}
    \caption{Example explanations for a model trained on original \gls{sst} (left) and models trained on decoy versions (right).
    For all sentences, groundtruth class and predicted class is `positive'.
    The input tokens are highlighted based on their contributions towards the prediction, from negative (red) to positive (green) contribution.
    % you sure green is low and red is high?
    We finetune \(\text{BERT}_{\text{base}}\)~\cite{devlin2019bert} with the default hyperparameter settings recommended in the original paper.
    The explainer is Integrated Gradients\footnotemark~\cite{sundararajan2017IntegratedGrads}.}
    
    \label{fig:explanation_examples}
\end{figure*}

We further formulate properties that decoys should satisfy for injecting \emph{right for the wrong reason} bugs:

\mpara{Adoptable.}
Discriminative machine learning models typically adopt the decision-rules offering the biggest reward w.r.t minimizing some objective function.
If there exists a simpler, more productive decision-rule than the one introduced by the decoy, a model might not adopt the latter and the decoy-rule is not learned.
While it is certainly possible to create decoy decision-rules based on complex natural language signals, we argue that a solution to the decoy should be either more \emph{superficial} or have a substantially higher \emph{productivity} than the solutions exposed by the original dataset.
Although the potential solutions to a dataset are typically not apparent to humans (otherwise one should probably refrain from using complex machine learning models), researchers and practitioners often have some intuition about the complexity of intended solutions to the task at hand.
The adoptability also depends on the decoy being representative. Its \emph{coverage} has to be reasonably high, such that it generalizes to a decent number of training instances.
Additionally, whether a decoy is adoptable depends on the inductive biases of the model, e.g.,~a decoy based on word positions is not adoptable by a bag-of-words model.

\mpara{Natural.}
Explanations are supposed to help detect right for the wrong reason bugs, which are difficult to identify from observing predictions and raw data alone.
It should be possible for a decoy to occur naturally, such that insights from evaluations on decoy datasets can potentially transfer to real-world scenarios.
A natural decoy also ensures that humans are not able to easily spot the decoy by observing raw data examples, which would defeat the purpose of using explanations in the first place.
Assuming the original dataset is \emph{natural}, the decoy dataset should adhere to its properties and distribution, at least on a per-instance level.
For example, for text tasks, the instances affected by a decoy should not violate grammar, syntax, or other linguistic properties, if these are also not violated in the original dataset.

% Another important aspect is how natural a decoy fits into the original dataset.
% The instances modified by adding a decoy should adhere to the original dataset's distribution as close as possible.
% For example, for text tasks, the instances affected by a decoy should not violate grammar, syntax or other linguistic properties, if they are also not violated in the original dataset.
% A natural decoy also ensures that humans are not able to easily spot the decoy by observing raw data examples, which would defeat the purpose of using explanations in the first place.
% This means that it has to be reasonably difficult for humans to differentiate between instances from the original dataset and the decoy dataset.

% \subsection{Anecdotal Example}

The first example in ~\cref{fig:explanation_examples} shows an explanation generated for a model trained on a decoy dataset corresponding to the first decoy variant of \gls{sst} used by \citet{Rieger2020InterpretationsAreUseful}.
In this decoy dataset, two class-indicator words are added at a random location in each sentence, with `text' indicating the positive class and `video' indicating the negative class.
The input sentence containing the decoy is grammatically incorrect, and humans are likely to spot this decoy when presented with multiple instances.
Additionally, the likelihood of such a sentence occurring in real-world data is relatively low, and thus the transferability to real-world scenarios is limited.

A more natural decoy is shown in rows 2 - 5 in ~\cref{fig:explanation_examples}, where we create a decoy dataset by removing all instances which contain the word `movie' and are labeled `negative', retaining the original dataset's naturalness on a local level.
Considering all test set instances containing the word `movie', the performance of a model trained on this decoy dataset drops to random chance (47.5\%), indicating that the model was indeed misled by the decoy rule even though its applicability is below 3.3\%.

\footnotetext{As provided by Captum~\cite{kokhlikyan2020captum}.}

\section{Efficient Debugging with Explanations}%Selecting instances for human examination.} % Which instances to explain?
\label{sec:efficiency_problem}
Another crucial component in the evaluation setups described in \cref{sec:eval_setups} is the choice of instances shown to the human subjects.
Such a selection is especially important when dealing with large datasets where the majority of instances have correct predictions with explanations aligning with human understanding.
Showing all instances to humans in order to isolate a few errors is inefficient and often infeasible as the inspection of many individual explanations is expensive in time and resources, especially when requiring domain experts.
Thus, the examination is typically conducted under a tight budget on the number of instances.

Apart from the greedy \gls{sp} algorithm proposed by \citet{ribeiro2016LIME}, this problem has been mostly brushed aside by assuming the selection process to be optimal.
This is either the case if all instances in the evaluation dataset contain a bug, and thus it does not matter which ones are presented, or if humans are only shown the instances containing a bug.
This assumption is problematic since it does not transfer to real-world scenarios where \emph{right for the wrong reasons} bugs often only apply to small minorities of instances.
Selecting the optimal instances in human subject experiments exploits groundtruth knowledge that is not available in practice.
For example, when inspecting the instances corresponding to rows 2 and 3 from \cref{fig:explanation_examples}, the `movie' bug is easily noticeable, while it is undetectable by observing rows 4 and 5.
When sampling instances of this decoy dataset uniformly, there is a chance of less than 3.3\% of being presented with an instance containing the bug.

As a result, an evaluation that assumes the selection process to be optimal might not reflect the actual utility of explanations for debugging in practical applications at all. Summarizing explanations, for example by spectral relevance clustering \cite{lapuschkin2019unmasking}, looks to be a promising way to boost the utility of explanations for tasks like debugging.

% \begin{itemize}
%     \item assumption in eval setups: selection process is optimal. this is the case if either all instances contain bug, or humans are only shown a selection of instances containing a bug.
%     \item in real world scenarios, this is not the case. the choice of instances to inspect is crucial. human examination is expensive and infeasible for large test sets.
%     \item from decoySST example: what if coverage is so low, that humans never see bug.
%     \item lack of work on this! only submodular pick algorithm by lime.
%     \item many metods that can help with this, e.g. using representative instances retrieved by clustering input data or explanations (spray), or prototypes and criticisms (MMD-critic)
% \end{itemize}

\section{Outlook}
Although the current evaluation setups provide a solid foundation, measuring the actual utility of explanations for debugging remains difficult and current evaluations might not transfer to real-world scenarios.
We envision a benchmark collection of carefully designed decoy datasets and buggy models to alleviate key limitations and accelerate the future development of new, utility-driven explanation methods, as well as methods improving the efficiency of current explanation techniques.

\section*{Acknowledgements}
We thank Schufa Holding AG for generously supporting this work.
Additionally, we thank the anonymous reviewers for their feedback.
\bibliography{references}
\bibliographystyle{acl_natbib}

% \appendix

% \section{Example Appendix}
% \label{sec:appendix}

% This is an appendix.

\end{document}